# Text Detoxification as Style Transfer in English and Hindi

**Sourabrata Mukherjee**[1], **Akanksha Bansal**[2], **Atul Kr. Ojha**[3,2],
**John P. McCrae**,[3] **Ondřej Dušek**[1]

[1]Charles University, Faculty of Mathematics and Physics, Prague, Czechia
[2]Panlingua Language Processing LLP, India
[3]Insight SFI Centre for Data Analytics, DSI, University of Galway, Ireland
{mukherjee,odusek}@ufal.mff.cuni.cz
akanksha.bansal@panlingua.co.in
{atulkumar.ojha,john.mccrae}@insight-centre.org

## Abstract

This paper focuses on text detoxification, i.e., automatically converting toxic text into non-toxic text. This task contributes to safer and more respectful online communication and can be considered a Text Style Transfer (TST) task, where the text's style changes while its content is preserved. We present three approaches: (i) knowledge transfer from a similar task (ii) multi-task learning approach, combining sequence-to-sequence modeling with various toxicity classification tasks, and (iii) delete and reconstruct approach. To support our research, we utilize a dataset provided by Dementieva et al. (2021), which contains multiple versions of detoxified texts corresponding to toxic texts. In our experiments, we selected the best variants through expert human annotators, creating a dataset where each toxic sentence is paired with a single, appropriate detoxified version. Additionally, we introduced a small Hindi parallel dataset, aligning with a part of the English dataset, suitable for evaluation purposes. Our results demonstrate that our approach effectively balances text detoxification while preserving the actual content and maintaining fluency.

**Content warning:** This paper contains examples that are toxic, offensive and/or sexist in nature.

## 1 Introduction

Text detoxification is a process that involves the automatic conversion of toxic text into non-toxic or detoxified text (Dale et al., 2021). Text detoxification can be viewed as a subtask within the broader context of Text Style Transfer (TST) (Dale et al., 2021). TST aims to alter the style of text while preserving its core content. In the case of text detoxification, the source style is toxic language, and the target style is non-toxic, with the primary objective being the transformation of text from a harmful and offensive nature to a neutral or positive one, all while retaining the rest of the original text's meaning.

Existing detoxification methods often rely on rule-based removal of toxic words or phrases (Dementieva et al., 2022), but this approach is not very efficient and can make sentences sound unnatural. It also does not consider whether the meaning of the sentence is affected by these removals. Additionally, because we have limited resources in terms of datasets (Dementieva et al., 2021) for transforming toxic text to non-toxic text, simple sequence-to-sequence training may not be enough for better results (Mukherjee and Dusek, 2023).

To improve this, we propose three approaches: (i) knowledge transfer from a similar task (ii) multi-task learning approach, utilizing sequence-to-sequence modeling coupled with various toxic and civil text classification tasks, and (iii) delete and reconstruct approach. These approaches enhance the sequence-to-sequence transformation and lead to better outcomes compared to basic sequence-to-sequence training.

To conduct our research and experiments, we leverage a dataset provided by Dementieva et al. (2021). This dataset contains various versions of text that have undergone detoxification, offering valuable resources for our investigations. We have taken a meticulous approach to ensure dataset quality. This includes soliciting the expertise of human annotators to select the most suitable detoxified versions for each toxic text, resulting in a carefully curated dataset where each source toxic sentence is paired with its corresponding non-toxic text.

Furthermore, to contribute to the development of multilingual parallel datasets for text detoxification, we present a novel dataset containing 500 parallel toxic and non-toxic sentences in Hindi, aligned with their English counterparts (see Section 3). Hindi is one of the 22 scheduled and official Indian languages and the largest speech community in India (Jha, 2010). This dataset serves as a validation resource for research in Hindi and will open the possibility of extending it to other Indian languages as well as in multilingual domains.

Our contributions are summarized as follows:

(i) We build on an existing English detoxified dataset. We thoughtfully curated the data with human experts, where each toxic sentence is matched with a single detoxified counterpart.

(ii) We introduce a novel dataset of 500 parallel sentences in Hindi aligned with their English counterparts for validation.

(iii) Our methodologies enhance text detoxification in low-resource settings through knowledge transfer, multitask learning, and delete and reconstruct mechanisms.

(iv) We achieve comparable results with external benchmarks.

(v) Our data and experimental code are released on GitHub.[1]

## 2 Related Work

**Text Style Transfer (TST)** Text Style Transfer (TST) is a task where texts with different styles but similar content are transformed (Hu et al., 2022). For instance, Jhamtani et al. (2017) used a model to turn modern English into Shakespearean English, and Mukherjee and Dusek (2023) explored TST with limited parallel data resources. However, TST is often tricky due to the shortage of matching data, with examples such as Yelp (sentiment) (Li et al., 2018) and the detoxification datasets of Dementieva et al. (2021, 2022) having the order of a few thousand examples at most. To tackle this, two main strategies have emerged: (i) simple text replacement, which involves explicitly finding and swapping specific style-related phrases (Li et al., 2018), (ii) implicit style separation, techniques like back-translation and autoencoding help separate style from content through hidden representations (Shen et al., 2017; Zhao et al., 2018; Fu et al., 2018; Prabhumoye et al., 2018; Hu et al., 2017).

---

[1]Code: https://github.com/souro/multilingual_text_detoxification, and data: https://github.com/panlingua/multilingual_text_detoxification_datasets.

**Toxicity** Several datasets have been built for the detection of toxic/hateful/offensive content from various social media platforms such as Twitter (now called *X*), Reddit, Facebook (Waseem and Hovy, 2016; Davidson et al., 2017; Kumar et al., 2018; Basile et al., 2019; Chakravarthi et al., 2021; Kumar et al., 2022; Kurrek et al., 2020; Bagga et al., 2021; Leonardelli et al., 2023; Kirk et al., 2023) and Wikipedia talk pages (Cjadams et al., 2017, 2019; Kivlichan et al., 2020). These datasets often contain various types of toxicity, including between offensive language and hateful/aggressive speech, whereas other datasets focus only on specific types of toxicity (Leonardelli et al., 2023; Kirk et al., 2023), e.g., sexism, religious discrimination, or racism. These datasets are explored to build automatic hate speech/offensive text detection models using various machine and deep learning approaches (Waseem and Hovy, 2016; Mandl et al., 2019; Zampieri et al., 2020; Chakravarthi et al., 2021; Kumar et al., 2022; Wiegand and Ruppenhofer, 2021). However, given that the focus is on toxic content detection and not detoxification, these datasets are not parallel (with aligned toxic and non-toxic sentence pairs), except for small data in English (Dementieva et al., 2021; Logacheva et al., 2022) and Russian (Dementieva et al., 2022), which leads to the same problems in detoxification as with TST in general (Hu et al., 2022).

To our best knowledge, the following models exist for the detoxification task: Duplicate, a rule-based Delete, RuT5, and RuPrompts baseline models (Dementieva et al., 2022). *Duplicate* keeps input text intact, establishing a trivial baseline. *Delete* is an unsupervised rule-based method that removes toxic words using a predefined vocabulary, akin to censoring on TV. *RuT5* involves supervised fine-tuning of the *T5* model. *RuPrompts* uses the ruPrompts library for continuous prompt tuning of language models (Konodyuk and Tikhonova, 2021).

## 3 Dataset Development

The original dataset (Dementieva et al., 2021) is a collection of user-generated comments which are toxic in nature. The features of the dataset can be summarised as follows:

(i) The utterances are a mix of miscellaneous domains ranging from political to personal to religious. Several sentences seem to be free of

context, that is, one cannot infer a specific domain/topic from its structure and content.

(ii) The source data is user-generated (comments and reviews) in real-time, hence it does not consist of well-formed sentences. Instead, the data consists of numerous typos, grammatical inaccuracies, and fragmented speech-like structures. After style transfer, the non-toxic/civil version remains error-free regarding typos. But the fragmented speech-like structure remains unchanged, as annotators must preserve the content.

### 3.1 Style Conversion Methodology

Based on observation of the original data (Dementieva et al., 2021), we are of the opinion that there are largely two ways utilized to transform toxic utterances into civil utterances:

(i) *Replacement* of abusive words or phrases with synonyms that soften the blow without changing the meaning. For example, *holy shit* can be expressed as *oh god, oh goodness* etc. In both cases, the element of excitement remains as-is, but the utterance sounds more civil. The problem with this approach is that one size does not fit all. Context and usage must be examined before choosing a suitable replacement.

(ii) Truncation: In certain cases, an abusive word or an abusive phrase is inserted in the utterance. Such phrases can be removed and the remaining sentence can be retained. For example, in the utterance *shit, I might have it.* simply removing *shit* makes the sentence non-toxic but retains the meaning.

### 3.2 Optimising the Original Dataset

In order to conduct our experiments and prepare for the Hindi translation, we performed several cleanup exercises:

(i) Of the 2,778 sentences in the data, only 1,108 are unique toxic sentences (Dementieva et al., 2021). The data comprised of 1-5 paraphrases for all the sentences. We manually selected the most suitable response for utterances with multiple variations available.

(ii) The data has been cleansed of numerical values. The authors have replaced numbers with *DIGIT*, *number*, or *<number>*. For example:

(a) *no fucking way number bricks fits in a car , this is bs* . (b) *DIGIT year olds can be little shits too, doesn t mean you fight them* . There was no way we could have located the original numbers and we did not create fictionalised numbers either. This feature has been carried forward in the translation as well.

(iii) There were certain foreign or non-English texts in the civil part of the database. We replaced such utterances with suitable English equivalents. For example, toxic: *dont be such a hypocrite*, civil: *não seja mentiroso*, corrected civil: *dont be such an unfair person*.

### 3.3 Toxic to non-Toxic parallel Hindi dataset

A random selection of 500 sentences was manually translated in Hindi. We kept the guidelines simple: maximum preservation of content and natural use of language. The following observations were made while translating this data.

(i) Direct and Implicit Toxicity: Most of the utterances contain direct toxicity with the use of abusive words. Implicit toxicity can also be seen in the dataset where the actions and wishes of the user make it toxic. For example, कैंसर पीड़ित हो जाओ और जितनी जल्दी हो सके मर जाओ। *(catch cancer and die as soon as possible)*. Here the wish for someone to die after contracting a fatal disease is implicitly toxic whereas explicit abusive words are lacking.

(ii) Choice of pronouns: Hindi pronouns can bear the weight of politeness and impoliteness and this linguistic feature has been used to make style transfer more pronounced. The toxic sentences, wherever possible, use the pronoun तू *(tu)*, whereas non-toxic would use आप *(aap)*.

(iii) Lack of style transfer: In certain cases, there is hardly any difference between toxic and civil due to the kind of abusive language used. For example, there is not much difference between *awful* and *god awful* in English, and the difference is lost completely when translated into Hindi like भयंकर *(bhayankar)* or भयानक *(bhayanak)*.

A few more such examples are documented in Table 1.

| ID | Toxic | Non-Toxic (Civil) |
|----|-------|-------------------|
| 1 | build a fucking coal plant there and let them all choke to death on their own crapulence . वहां साला एक कोयला प्लांट बनाओ और उन सभी को अपनी खुद की बकचोदी से घुट कर मारने दो। | assemble a screwing coal plant there and let them all stifle to death on their own crapulence वहां साला एक कोयला प्लांट बनाओ और उन सभी को अपनी खुद की बकवास से घुट कर मारने दो। |
| 2 | move to europe if you want to live with stupid . अगर तू बेवकूफ के साथ रहना चाहता है तो यूरोप चला जा। | I don't think it is a good idea to move to Europe. मुझे नहीं लगता कि यूरोप में जाना एक अच्छा विचार है। |
| 3 | ya unlike the dems who haven t done shit about climate change. हां, उन डेम्स के विपरीत जिन्होंने जलवायु परिवर्तन के बारे में घंटा कुछ नहीं किया है। | ya unlike the dems who haven t done anything about climate change . हां, उन डेम्स के विपरीत जिन्होंने जलवायु परिवर्तन के बारे में कुछ नहीं किया है। |
| 4 | thats a great fucking point . एक बहुत ही कमाल का तर्क है। | That is a fantastic point. यह एक शानदार तर्क है। |
| 5 | go fuck yourself , and not in a kindly matter at all . अपनी गांड मार ले, और हल्के से मत मारियो। | mind your own stuff अपने काम से काम रख। |

Table 1: A set of examples from English and Hindi detoxification dataset.

## 4 Detoxification Models

We first introduce the seq2seq baseline (*seq2seq*, Section 4.1), then present three distinct extension approaches: Knowledge Transfer from a Similar Task (*kt*, Section 4.2), Multitask Learning (Section 4.3), and Delete and Reconstruct (*del_recon*, Section 4.4) for the text detoxification task.

### 4.1 Sequence-to-Sequence Baseline *(seq2seq)*

The baseline is a sequence-to-sequence learning approach using mBART (Liu et al., 2020a) with parallel data. We use the cross-entropy loss for the sequence-to-sequence task, defined as:

$$L_{seq2seq} = -\sum_{t=1}^{T}\sum_{k=1}^{K} \log(P(y_{t,k}|X)), \quad (1)$$

where $L_{seq2seq}$ represents the sequence-to-sequence loss, $X$ is the input text, $T$ is the length of the sequence, $K$ is the vocabulary size, and $P(y_{t,k}|X)$ is the predicted probability of the $k$-th token at time step $t$ given the input $X$.

### 4.2 Knowledge Transfer from a Similar Task *(kt)*

In scenarios with limited resources, leveraging knowledge from a related task can enhance our approach. To achieve this, we employ a two-step process. First, we fine-tune a model to perform the negative-to-positive text sentiment transfer task using the text sentiment transfer yelp dataset provided by Li et al. (2018). Subsequently, we transfer the acquired knowledge in the form of model weights to further fine-tune the model using our toxic-to-non-toxic data, in the same fashion as the *seq2seq* baseline.

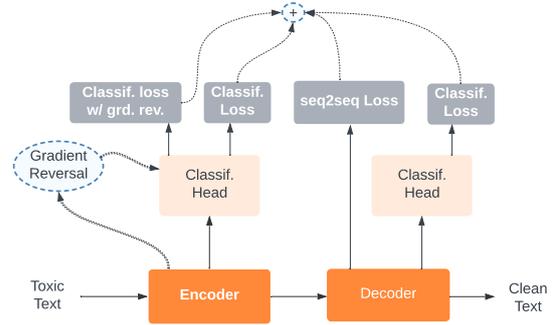

Figure 1: Overview of the Multitask Learning Methodology.

### 4.3 Multitask Learning

In this approach, we employ a multitask learning setup to transfer toxic attributes in text to non-toxic attributes. This involves learning multiple tasks simultaneously. For an overview of this methodology, see Figure 1.

We introduce several classification tasks, each of which works in conjunction with the primary sequence-to-sequence task (*seq2seq*, Section 4.1):

(i) **Classification of Input Text *(cls_ip)*:** This task aims to classify the input text (in the encoder) as toxic or non-toxic. The associated loss, combined with the sequence-to-sequence loss, is defined as:

$$L_{cls\_ip} = -\sum_{i=1}^{N}[t_i\log(P_{cls\_ip}(x_i)) + \\ (1-t_i)\log(1-P_{cls\_ip}(x_i))], \quad (2)$$

where $L_{cls\_ip}$ is the toxicity classification loss, $N$ is the number of training samples, $x_i$ is the input text sample, $t_i$ is the corresponding toxicity label (0 for toxic, 1 for non-toxic), and

$P_{cls\_ip}(x_i)$ is the predicted probability of toxicity for input $x_i$.

(ii) **Classification with a Gradient Reversal Layer** *(cls_gr_ip)*: Similar to the classification of input text, in addition, this task includes a gradient reversal layer before the classification head. The gradient reversal layer effectively scales the gradient flowing to the encoders by a factor of $-\lambda$, which should help keep representations of similar-meaning toxic and non-toxic texts similar, focusing on content preservation.

$$Grad_{rev} = -\lambda \cdot \nabla J \quad (3)$$

where $-\lambda$ represents a scaling factor that multiplies the gradient. $\nabla J$ (grad_output) is the gradient flowing through the network.

Then the classification loss is defined as:

$$L_{cls\_gr\_ip} = -\sum_{i=1}^{N}[t_i \log(P_{cls\_gr\_ip}(x_i)) \\ + (1-t_i)\log(1-P_{cls\_gr\_ip}(x_i))], \quad (4)$$

where $L_{cls\_gr\_ip}$ is the classification with gradient reversal layer, and $P_{cls\_gr\_ip}(x_i)$ is the predicted probability after applying the gradient reversal layer (Equation 3).

(iii) **Classification of Generated Output Text** *(cls_op)*: This task focuses on detecting whether the generated output text (in the decoder) is toxic or non-toxic. The loss is defined similarly to the previous classification tasks:

$$L_{cls\_op} = -\sum_{i=1}^{N}[d_i \log(P_{cls\_op}(y_i)) + \\ (1-d_i)\log(1-P_{cls\_op}(y_i))], \quad (5)$$

where $L_{cls\_op}$ is the loss for detecting generated output text toxic or non-toxic, $y_i$ is the generated output text sample, $d_i$ is the corresponding target toxicity label (0 for toxic, 1 for non-toxic), and $P_{cls\_op}(y_i)$ is the predicted probability of non-toxicity.

### 4.4 Delete and Reconstruct *(del_recon)*

This approach is shown in Figure 2. We start with a toxicity classifier trained to differentiate between toxic (1) and non-toxic (0) sentences, using the training portion of our dataset (see Section 6). Leveraging this same classifier, we calculate word attributions for all sentences, encompassing both toxic and non-toxic examples. We then selectively remove words with attributions exceeding a threshold of 0.5.

| Language | Classifiers Accuracy (%) |
|---|---|
| English | 91.7 |
| Hindi | 59.8 |

Table 2: English and Hindi accuracy scores for Toxicity classifiers.

In the training phase, we fed the sentences after eliminating toxic words or phrases into mBART (Liu et al., 2020b) with non-toxic text from the dataset serving as the target output (see Equation 6):

$$L_{\text{reconstruction}} = \sum_{i=1}^{N}[y_i \log(P_{\text{reconstruction}}(x_i)) \\ + (1-y_i)\log(1-P_{\text{reconstruction}}(x_i))] \quad (6)$$

The ultimate goal is to preserve non-toxic content while generating natural and clean text through this process. The loss measures the difference between the reconstructed and original sentences. In Equation 6), $L_{\text{reconstruction}}$ is the reconstruction loss, $P_{\text{reconstruction}}(x_i)$ is the predicted probability of the reconstructed sentence, $y_i$ is the original non-toxic sentence, and $x_i$ represents the input sentence (where toxic words have been deleted).

## 5 Experimental Settings

To ensure consistency in our experiments, we partitioned the English datasets into 508 examples for training, 100 for development, and 500 for testing. For the Hindi dataset, we created training and development sets of the same size as the English dataset through machine translation. We utilized the Facebook NLLB-200-3.3B model (Costa-jussà et al., 2022) available from HuggingFace. For evaluation in Hindi, we employed our newly provided dataset consisting of 500 samples (see Section 3). We employed the mBART-large-50 model (Tang et al., 2020) from the HuggingFace library (Wolf et al., 2020) for both English and Hindi. To optimize model performance, we conducted hyperparameter tuning, leading to the selection of a learning rate of 1e-5 and a batch size of 3. Throughout the network, dropout was applied with a rate of 0.1.

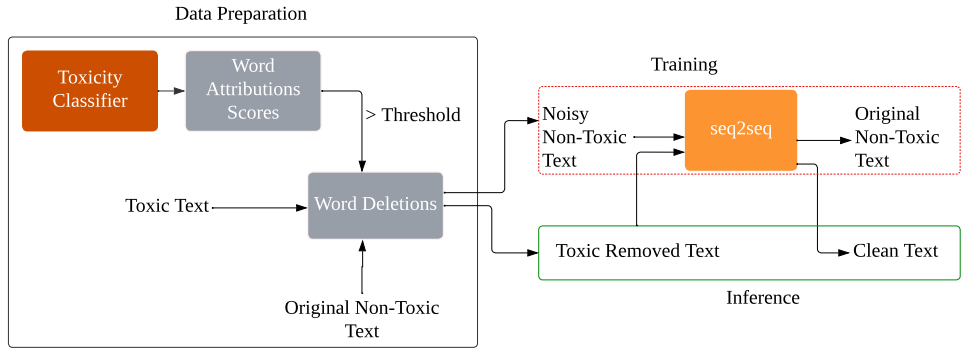

Figure 2: Overview of the Delete and Reconstruct Methodology.

Additionally, we introduced L2 regularization with a strength of 0.01. The training was executed over 5 epochs.

## 6 Evaluation Metrics

The evaluation process involves three primary aspects: accuracy of the toxic to civil text transfer (detoxification accuracy), content preservation, and fluency. Detoxification accuracy is assessed using our fine-tuned mBERT classifier, which used the same training set for finetuning as our primary TST task (see Section 3). Classifier accuracies of toxic and non-toxic text in English and Hindi languages are shown in Table 2. The rather low accuracy in Hindi might be a result of the fact that the classifier is finetuned on synthetic training and development sets created by machine English-to-Hindi translation, while it is evaluated using manually translated data.[2] Content preservation is evaluated through BLEU score (Papineni et al., 2002) and embedding similarity (Rahutomo et al., 2012) compared against the input sentences, where embedding similarity is determined using language-agnostic BERT sentence embeddings (LaBSE) (Feng et al., 2022) in conjunction with cosine similarity. Evaluating fluency, particularly for Hindi, poses a challenge due to the limited availability of assessment tools for Indic languages (Krishna et al., 2022). Although perplexity (PPL) tends to favor unnatural sentences with common words and may not be ideal for fluency evaluation (Pang, 2019; Mir et al., 2019), we include a basic fluency assessment using perplexity (PPL) measured with a multilingual GPT model (Shliazhko et al., 2022).

As automated metrics for language generation may not correlate well with human judgments (Novikova et al., 2017), we also run a small-scale human evaluation with language expert annotators on a random sample of 50 sentences from the test set for each language. Outputs are rated on a 5-point Likert scale for detoxification accuracy, content preservation, and fluency.

## 7 External Baselines

Dementieva et al. (2022) provided two of their *RuT5* and *Delete* detoxification baseline methods publicly. We could not use them directly for a result comparison as they are only designed for the Russian language. Therefore, we adapted the *RuT5* model,[3] which is based on the Russian language, using *t5-base* (Raffel et al., 2020) for English and mt5-small (Xue et al., 2021) for Hindi. For the *Delete* method, Dementieva et al. used a dictionary of toxic words and/or phrases. To generate non-toxic sentences, they simply deleted from toxic sentences all toxic words and phrases contained in their dictionary. To adopt this method, we translated their dictionary from Russian to English and then English to Hindi using *Google Translate*[4] and then applied the same technique.

## 8 Results and Analysis

### 8.1 Automatic Evaluation

Automatic evaluation results are presented in Table 3.

**Performance of Our Methodologies:** The *seq2seq* baseline model showed moderate performance across all metrics, indicating its basic capability in text detoxification. Our knowledge transfer methodology (*kt*) exhibited a substantial improvement in text detoxification accuracy and content preservation over the baseline, thanks to the knowledge transferred from a similar task. In the Multitask Learning setup, *seq2seq + cls_ip* model achieved average results in text detoxification accuracy, content preservation, and fluency. The *seq2seq + cls_gr_ip* approach displayed exceptional text detoxification accuracy but at the cost of fluency and content preservation. The *seq2seq + cls_op* model demonstrated good overall performance with balanced results in text detoxification accuracy, content preservation, and fluency. Our *del_recon* methodology also demonstrates good performance in terms of detoxification accuracy (ACC) and content preservation scores (BLEU and CS), which are the highest overall, but at the cost of poor fluency.

---

[2]We observed that some of the Hindi machine translation outputs for English toxic inputs are less toxic or not toxic at all. While this may be a generally desired outcome of machine translation, it makes our task of achieving clear toxic and non-toxic classification more challenging.

[3]https://huggingface.co/ai-forever/ruT5-base
[4]https://translate.google.com/

|  | **English** | | | | **Hindi** | | | |
|---|---|---|---|---|---|---|---|---|
| **Models** | ACC | BLEU | CS | PPL | ACC | BLEU | CS | PPL |
| *Our Baseline* | | | | | | | | |
| seq2seq | 67.4 | 43.1 | 76.8 | 221.4 | 68.4 | 39.6 | 77.2 | 8.5 |
| *Our Methodology - Knowledge Transfer* | | | | | | | | |
| kt | 71.0 | 45.6 | 77.5 | 237.9 | 92.0 | 42.0 | 78.6 | 9.3 |
| *Our Methodology - Multitask Learning* | | | | | | | | |
| seq2seq + cls_ip | 64.0 | 43.7 | 75.6 | 202.4 | 77.2 | 38.5 | 76.8 | 8.3 |
| seq2seq + cls_gr_ip | 95.6 | 0.2 | 16.3 | 20.4 | 75.2 | 36.2 | 72.6 | 8.2 |
| seq2seq + cls_op | 75.8 | 44.2 | 76.6 | 348.3 | 79.8 | 39.8 | 78.2 | 8.2 |
| *Our Methodology - Delete and Reconstruct* | | | | | | | | |
| del_recon | 80.6 | 44.5 | 76.9 | 304.6 | 94.0 | 41.2 | 78.9 | 8.2 |
| *External Baselines (see Section 7)* | | | | | | | | |
| Delete | 68.6 | 41.0 | 74.4 | 599.7 | 92.8 | 40.4 | 76.8 | 11.4 |
| T5 | 59.2 | 44.7 | 77.1 | 221.6 | 99.6 | 1.2 | 38.7 | 64.7 |

Table 3: Automatic evaluation results. We measure the detoxification accuracy (ACC) using a toxicity classifier, BLEU score, Content Similarity (CS), and fluency using perplexity (PPL) (see Section 6). Model names follow the conventions introduced in Section 4.

**Language-wise Analysis:** The results in English and Hindi exhibit mostly the same trends, with methodologies that perform well in one language tending to perform well in the other. However, it is worth noting that while our models maintain mostly good performance in both languages, English text detoxification consistently demonstrates slightly better results. The performance in Hindi is a little inconsistent, possibly due to the use of synthetic data for training (cf. Section 6).

**Comparison with External Baselines:** Comparing our methodologies with external baselines, it is evident that our models outperform both the *Delete* and *T5* baselines in most aspects. Our models generally exhibit more balanced results across multiple metrics, demonstrating their effectiveness in addressing text detoxification challenges.

In summary, our knowledge transfer methodology (*kt*) exhibits notable advancements in text detoxification, with balanced results across different metrics. The multi-task learning approaches show promise, with variations that excel in specific aspects. These findings underscore the potential of our methodologies to enhance text detoxification tasks, fostering safer and more respectful online communication.

### 8.2 Human Evaluation

A group of experts performed the human evaluation exercise, for which we chose the following four models: (1) *seq2seq+cls_op* as one of our best models from the multitask learning experiment, (2) *kt*, (3) *del_recon*, and (4) *Delete* as the best external baseline. The choice was based on the overall balanced results on automatic metrics (see Section 8.1. The results, shown in Table 4, mostly align with our automatic evaluation findings, validating the effectiveness of the data and our proposed methodologies.

### 8.3 Sample Output

The sample outputs in Table 5 provide an overview of the performance of all the systems used in the human evaluation across English and Hindi.

The *Delete* baseline's output is inaccurate as compared to *seq2seq + cls_op*. The *Delete* base-line's output deletes the abusive lexical units, making the sentence syntactically incorrect. If we look closely at the examples presented here, we notice that the adjective or the adverb previously attached to an abusive noun remains syntactically disconnected, as the noun has not been replaced with any other non-abusive lexical unit. Similarly, when we compare the outputs of *kt* and *del_recon*, we find that the former replaces the abusive word with another noun in most cases, whereas *del_recon* simply truncates the abusive word, creating a syntactic inconsistency in the sentence. The results for Hindi are not on par with English for any model.

| Models | English | | | Hindi | | |
|---|---|---|---|---|---|---|
| | Accuracy | Content | Fluency | Accuracy | Content | Fluency |
| *Our Methodologies* | | | | | | |
| kt | 3.0 | 4.9 | 4.9 | 1.9 | 4.8 | 4.8 |
| seq2seq + cls_op | 3.2 | 4.9 | 4.8 | 2.0 | 4.9 | 4.5 |
| del_recon | 3.3 | 4.7 | 3.9 | 2.0 | 4.9 | 4.7 |
| *External Baseline* | | | | | | |
| Delete | 2.6 | 4.9 | 3.0 | 1.9 | 4.9 | 3.9 |

Table 4: Human evaluation of 50 randomly selected outputs on toxic to non-toxic transfer accuracy (Accuracy), Content Preservation (Content), and Fluency (see Section 8.2).

## 9 Conclusion

We explored the task of text detoxification, aiming to automatically transform toxic text into non-toxic text while preserving content and fluency. Our findings suggest that our approaches successfully address the task of text detoxification, achieving notable detoxification accuracy while preserving content and maintaining fluency. In future research, we will focus on exploring text detoxification in various linguistic and cultural contexts. We anticipate that our work will pave the way for more effective and ethical online communication by mitigating the harmful impact of toxic text.

## Limitations

While conducting our study, we identified certain limitations that warrant consideration. Firstly, the efficacy of our approach may vary across different languages and styles of toxicity. The nuances in linguistic expression and cultural context pose challenges for creating a one-size-fits-all solution. Further research is essential to adapt and enhance our methods to cater to the diverse linguistic landscape of online content.

Additionally, we acknowledge the difficulty in fluency evaluation. The absence of precise assessment tools makes it challenging to measure how well our text detoxification process maintains the natural flow and coherence of the original content. This limitation underscores the need for the development of robust evaluation metrics to ensure the overall quality of the detoxified text.

Recognizing these limitations, we emphasize the importance of continued research to address these challenges systematically. Refining our methods through iterative improvements and adapting to the evolving nature of online communication will be crucial in achieving broader effectiveness and applicability.

## Ethics Statement

Our commitment to ethical practices is at the core of our research endeavors. The text detoxification process we employ prioritizes principles of respect and safety, with the overarching goal of reducing harmful content while preserving the authentic meaning of the text. We are acutely aware of the significance of cultivating a positive online environment.

In our pursuit of responsible and unbiased AI development, we strive to minimize any unintended consequences or biases in our work. This commitment involves continuous scrutiny of our methods and dedication to transparency in communicating the potential limitations and ethical considerations associated with our approach.

Moreover, we recognize the dynamic nature of ethical standards in AI and are committed to staying abreast of evolving ethical guidelines. Our dedication to ethical conduct extends beyond the immediate scope of this research, influencing our broader vision for the positive impact of AI on society.

| Source → Target | kt (Ours) | seq2seq + cls_op (Ours) | del_recon (Ours) | Delete (External baseline) |
|---|---|---|---|---|
| and telling nato to fuck off . → and telling nato to leave. और नाटो को चलता करना। → और नाटो को जाने के लिए कहना। | and telling nato to stop. नाटो को चलता करना | and telling nato to stop और नाटो को चलता करना. | and telling nato to off. और नाटो को चलता करना। | and telling nato to off . और नाटो को चलता करना। |
| ours is pretty fucked up too . → ours isnt good either. हमारे वाले की भी भयंकर लगी पड़ी है। → हमारा वाला भी अच्छा नहीं है। | ours is pretty down too हमारे वाले की भी भयंकर लगी है। | ours is pretty bad too हमारे वाले की भी भयंकर लगी पड़ी है। | ours is pretty up too. हमारे वाले की भी भयंकर लगी पड़ी है। | ours is up too . हमारे वाले की भी भयंकर लगी पड़ी है। |
| oh shit you just blew their minds . → Oh you just blew their minds तेरी माँ की आँख, तूने उनके दिमाग़ की दही कर दी। → ओह, तुमने उनको अचंभित कर दिया। | oh shit you just blew their minds तेरी माँ की आँख, उनके दिमाग़ की दही कर दी। | Oh no you just blew their minds तेरी माँ की आँख, तूने उनके दिमाग़ की दही कर दी। | oh you just blew their minds. तूने उनके दिमाग़ की दही कर दी। | oh you just blew their minds . तेरी माँ की आँख, तूने उनके दिमाग़ की दही कर दी। |
| he is an autocratic bully . → he is a bully वह एक निरंकुश धमकाने वाला है। → वह एक बदमाश है | he is an autocratic person वह एक निरंकुश धमकाने वाला है | He is not good वह एक निरंकुश धमकाने वाला है | he is an arrogant leader. वह एक निरंकुश धमकाने वाला है। | he is an autocratic . वह एक धमकाने वाला है। |
| you are a veritable idiot ! → I disagree with you तू एक सत्यनिष्ठ बेवकूफ है! → मैं आप से असहमत हूं | You are unbelievable! तू एक सत्यनिष्ठ व्यक्ति है! | You are not good आप एक सत्यनिष्ठ व्यक्ति है! | you are a veritable person तू एक है? | you are a veritable ! तू एक सत्यनिष्ठ व्यक्ति है! |

Table 5: Selective Model's (see Section 8.3) sample outputs in English and Hindi.

## Acknowledgements

This research was funded by the European Union (ERC, NG-NLG, 101039303) and by Charles University projects GAUK 392221 and SVV 260698. We acknowledge the use of resources provided by the LINDAT/CLARIAH-CZ Research Infrastructure (Czech Ministry of Education, Youth, and Sports project No. LM2018101). We also acknowledge Panlingua Language Processing LLP for providing the dataset and this collaborative research project.

Atul Kr. Ojha and John P. McCrae would like to acknowledge the support of the Science Foundation Ireland (SFI) as part of Grant Number SFI/12/RC/2289_P2 Insight_2, Insight SFI Centre for Data Analytics.

## References

Sunyam Bagga, Andrew Piper, and Derek Ruths. 2021. "are you kidding me?": Detecting unpalatable questions on Reddit. In *Proceedings of the 16th Conference of the European Chapter of the Association for Computational Linguistics: Main Volume*, pages 2083–2099, Online. Association for Computational Linguistics.

Valerio Basile, Cristina Bosco, Elisabetta Fersini, Debora Nozza, Viviana Patti, Francisco Manuel Rangel Pardo, Paolo Rosso, and Manuela Sanguinetti. 2019. SemEval-2019 task 5: Multilingual detection of hate speech against immigrants and women in Twitter. In *Proceedings of the 13th International Workshop on Semantic Evaluation*, pages 54–63, Minneapolis, Minnesota, USA. Association for Computational Linguistics.

Bharathi Raja Chakravarthi, Ruba Priyadharshini, Navya Jose, Anand Kumar M, Thomas Mandl, Prasanna Kumar Kumaresan, Rahul Ponnusamy, Hariharan R L, John P. McCrae, and Elizabeth Sherly. 2021. Findings of the shared task on offensive language identification in Tamil, Malayalam, and Kannada. In *Proceedings of the First Workshop on Speech and Language Technologies for Dravidian Languages*, pages 133–145, Kyiv. Association for Computational Linguistics.

Cjadams, Daniel Borkan, inversion, Jeffrey Sorensen, Lucas Dixon, Lucy Vasserman, and Nithum. 2019. Jigsaw Unintended Bias in Toxicity Classification.

Cjadams, Jeffrey Sorensen, Lucas Dixon Julia Elliott, Mark McDonald, Nithum, and Will Cukierski. 2017. Toxic Comment Classification Challenge.

Marta R. Costa-jussà, James Cross, Onur Çelebi, Maha Elbayad, Kenneth Heafield, Kevin Heffernan, Elahe Kalbassi, Janice Lam, Daniel Licht, Jean Maillard, Anna Sun, Skyler Wang, Guillaume Wenzek, Al Youngblood, Bapi Akula, Loïc Barrault, Gabriel Mejia Gonzalez, Prangthip Hansanti, John Hoffman, Semarley Jarrett, Kaushik Ram Sadagopan, Dirk Rowe, Shannon Spruit, Chau Tran, Pierre Andrews, Necip Fazil Ayan, Shruti Bhosale, Sergey Edunov, Angela Fan, Cynthia Gao, Vedanuj Goswami, Francisco Guzmán, Philipp Koehn, Alexandre Mourachko, Christophe Ropers, Safiyyah Saleem, Holger Schwenk, and Jeff

Wang. 2022. No language left behind: Scaling human-centered machine translation. *CoRR*, abs/2207.04672.

David Dale, Anton Voronov, Daryna Dementieva, Varvara Logacheva, Olga Kozlova, Nikita Semenov, and Alexander Panchenko. 2021. Text detoxification using large pre-trained neural models. In *Proceedings of the 2021 Conference on Empirical Methods in Natural Language Processing*, pages 7979–7996, Online and Punta Cana, Dominican Republic. Association for Computational Linguistics.

Thomas Davidson, Dana Warmsley, Michael W. Macy, and Ingmar Weber. 2017. Automated hate speech detection and the problem of offensive language. In *Proceedings of the Eleventh International Conference on Web and Social Media, ICWSM 2017, Montréal, Québec, Canada, May 15-18, 2017*, pages 512–515. AAAI Press.

Daryna Dementieva, Varvara Logacheva, Irina Nikishina, Alena Fenogenova, David Dale, Irina Krotova, Nikita Semenov, Tatiana Shavrina, and Alexander Panchenko. 2022. Russe-2022: Findings of the first russian detoxification shared task based on parallel corpora. volume 2022, page 114 – 131. Cited by: 1; All Open Access, Bronze Open Access.

Daryna Dementieva, Sergey Ustyantsev, David Dale, Olga Kozlova, Nikita Semenov, Alexander Panchenko, and Varvara Logacheva. 2021. Crowdsourcing of parallel corpora: the case of style transfer for detoxification. In *Proceedings of the 2nd Crowd Science Workshop: Trust, Ethics, and Excellence in Crowdsourced Data Management at Scale co-located with 47th International Conference on Very Large Data Bases (VLDB 2021), Copenhagen, Denmark, August 20, 2021*, volume 2932 of *CEUR Workshop Proceedings*, pages 35–49. CEUR-WS.org.

Fangxiaoyu Feng, Yinfei Yang, Daniel Cer, Naveen Arivazhagan, and Wei Wang. 2022. Language-agnostic BERT sentence embedding. In *Proceedings of the 60th Annual Meeting of the Association for Computational Linguistics (Volume 1: Long Papers), ACL 2022, Dublin, Ireland, May 22-27, 2022*, pages 878–891. Association for Computational Linguistics.

Zhenxin Fu, Xiaoye Tan, Nanyun Peng, Dongyan Zhao, and Rui Yan. 2018. Style transfer in text: Exploration and evaluation. In *Proceedings of the Thirty-Second AAAI Conference on Artificial Intelligence, (AAAI-18), the 30th innovative Applications of Artificial Intelligence (IAAI-18), and the 8th AAAI Symposium on Educational Advances in Artificial Intelligence (EAAI-18), New Orleans, Louisiana, USA, February 2-7, 2018*, pages 663–670. AAAI Press.

Zhiqiang Hu, Roy Ka-Wei Lee, Charu C. Aggarwal, and Aston Zhang. 2022. Text style transfer: A review and experimental evaluation. *SIGKDD Explor.*, 24(1):14–45.

Zhiting Hu, Zichao Yang, Xiaodan Liang, Ruslan Salakhutdinov, and Eric P. Xing. 2017. Toward controlled generation of text. In *Proceedings of the 34th International Conference on Machine Learning, ICML 2017*, volume 70 of *Proceedings of Machine Learning Research*, pages 1587–1596, Sydney, NSW, Australia. PMLR.

Girish Nath Jha. 2010. The TDIL program and the Indian langauge corpora intitiative (ILCI). In *Proceedings of the Seventh International Conference on Language Resources and Evaluation (LREC'10)*, Valletta, Malta. European Language Resources Association (ELRA).

Harsh Jhamtani, Varun Gangal, Eduard Hovy, and Eric Nyberg. 2017. Shakespearizing modern language using copy-enriched sequence to sequence models. In *Proceedings of the Workshop on Stylistic Variation*, pages 10–19, Copenhagen, Denmark. Association for Computational Linguistics.

Hannah Kirk, Wenjie Yin, Bertie Vidgen, and Paul Röttger. 2023. SemEval-2023 task 10: Explainable detection of online sexism. In *Proceedings of the 17th International Workshop on Semantic Evaluation (SemEval-2023)*, pages 2193–2210, Toronto, Canada. Association for Computational Linguistics.

Ian Kivlichan, Jeffrey Sorensen, Julia Elliott, Lucy Vasserman, Martin Görner, and Phil Culliton. 2020. Jigsaw Multilingual Toxic Comment Classification.

Nikita Konodyuk and Maria Tikhonova. 2021. Continuous prompt tuning for russian: How to learn prompts efficiently with rugpt3? In *Recent Trends in Analysis of Images, Social Networks and Texts - 10th International Conference, AIST 2021, Tbilisi, Georgia, December 16-18, 2021, Revised Supplementary Proceedings*, volume 1573 of *Communications in Computer and Information Science*, pages 30–40. Springer.

Kalpesh Krishna, Deepak Nathani, Xavier Garcia, Bidisha Samanta, and Partha Talukdar. 2022. Few-shot controllable style transfer for low-resource multilingual settings. In *Proceedings of the 60th Annual Meeting of the Association for Computational Linguistics (Volume 1: Long Papers), ACL 2022, Dublin, Ireland, May 22-27, 2022*, pages 7439–7468. Association for Computational Linguistics.

Ritesh Kumar, Atul Kr. Ojha, Shervin Malmasi, and Marcos Zampieri. 2018. Benchmarking aggression identification in social media. In *Proceedings of the First Workshop on Trolling, Aggression and Cyberbullying (TRAC-2018)*, pages 1–11, Santa Fe, New Mexico, USA. Association for Computational Linguistics.

Ritesh Kumar, Shyam Ratan, Siddharth Singh, Enakshi Nandi, Laishram Niranjana Devi, Akash Bhagat, Yogesh Dawer, Bornini Lahiri, Akanksha Bansal, and Atul Kr. Ojha. 2022. The ComMA dataset v0.2:

Annotating aggression and bias in multilingual social media discourse. In *Proceedings of the Thirteenth Language Resources and Evaluation Conference*, pages 4149–4161, Marseille, France. European Language Resources Association.

Jana Kurrek, Haji Mohammad Saleem, and Derek Ruths. 2020. Towards a comprehensive taxonomy and large-scale annotated corpus for online slur usage. In *Proceedings of the Fourth Workshop on Online Abuse and Harms*, pages 138–149, Online. Association for Computational Linguistics.

Elisa Leonardelli, Gavin Abercrombie, Dina Almanea, Valerio Basile, Tommaso Fornaciari, Barbara Plank, Verena Rieser, Alexandra Uma, and Massimo Poesio. 2023. SemEval-2023 task 11: Learning with disagreements (LeWiDi). In *Proceedings of the 17th International Workshop on Semantic Evaluation (SemEval-2023)*, pages 2304–2318, Toronto, Canada. Association for Computational Linguistics.

Juncen Li, Robin Jia, He He, and Percy Liang. 2018. Delete, retrieve, generate: a simple approach to sentiment and style transfer. In *Proceedings of the 2018 Conference of the North American Chapter of the Association for Computational Linguistics: Human Language Technologies, NAACL-HLT 2018, Volume 1 (Long Papers)*, pages 1865–1874, New Orleans, Louisiana, USA. Association for Computational Linguistics.

Yinhan Liu, Jiatao Gu, Naman Goyal, Xian Li, Sergey Edunov, Marjan Ghazvininejad, Mike Lewis, and Luke Zettlemoyer. 2020a. Multilingual denoising pre-training for neural machine translation. *Transactions of the Association for Computational Linguistics*, 8:726–742.

Yinhan Liu, Jiatao Gu, Naman Goyal, Xian Li, Sergey Edunov, Marjan Ghazvininejad, Mike Lewis, and Luke Zettlemoyer. 2020b. Multilingual denoising pre-training for neural machine translation. *Trans. Assoc. Comput. Linguistics*, 8:726–742.

Varvara Logacheva, Daryna Dementieva, Sergey Ustyantsev, Daniil Moskovskiy, David Dale, Irina Krotova, Nikita Semenov, and Alexander Panchenko. 2022. ParaDetox: Detoxification with parallel data. In *Proceedings of the 60th Annual Meeting of the Association for Computational Linguistics (Volume 1: Long Papers)*, pages 6804–6818, Dublin, Ireland. Association for Computational Linguistics.

Thomas Mandl, Sandip Modha, Prasenjit Majumder, Daksh Patel, Mohana Dave, Chintak Mandalia, and Aditya Patel. 2019. Overview of the HASOC track at FIRE 2019: Hate speech and offensive content identification in Indo-European languages. In *FIRE '19: Forum for Information Retrieval Evaluation, Kolkata, India, December, 2019*, pages 14–17. ACM.

Remi Mir, Bjarke Felbo, Nick Obradovich, and Iyad Rahwan. 2019. Evaluating style transfer for text. In *Proceedings of the 2019 Conference of the North American Chapter of the Association for Computational Linguistics: Human Language Technologies, Volume 1 (Long and Short Papers)*, pages 495–504, Minneapolis, Minnesota. Association for Computational Linguistics.

Sourabrata Mukherjee and Ondrej Dusek. 2023. Leveraging low-resource parallel data for text style transfer. In *Proceedings of the 16th International Natural Language Generation Conference*, pages 388–395, Prague, Czechia. Association for Computational Linguistics.

Jekaterina Novikova, Ondřej Dušek, Amanda Cercas Curry, and Verena Rieser. 2017. Why we need new evaluation metrics for NLG. In *Proceedings of the 2017 Conference on Empirical Methods in Natural Language Processing*, pages 2241–2252, Copenhagen, Denmark. Association for Computational Linguistics.

Richard Yuanzhe Pang. 2019. Towards actual (not operational) textual style transfer auto-evaluation. In *Proceedings of the 5th Workshop on Noisy User-generated Text (W-NUT 2019)*, pages 444–445, Hong Kong, China. Association for Computational Linguistics.

Kishore Papineni, Salim Roukos, Todd Ward, and Wei-Jing Zhu. 2002. BLEU: a method for automatic evaluation of machine translation. In *Proceedings of the 40th Annual Meeting of the Association for Computational Linguistics*, pages 311–318, Philadelphia, Pennsylvania, USA. Association for Computational Linguistics.

Shrimai Prabhumoye, Yulia Tsvetkov, Ruslan Salakhutdinov, and Alan W Black. 2018. Style transfer through back-translation. In *Proceedings of the 56th Annual Meeting of the Association for Computational Linguistics (Volume 1: Long Papers)*, pages 866–876, Melbourne, Australia. Association for Computational Linguistics.

Colin Raffel, Noam Shazeer, Adam Roberts, Katherine Lee, Sharan Narang, Michael Matena, Yanqi Zhou, Wei Li, and Peter J. Liu. 2020. Exploring the limits of transfer learning with a unified text-to-text transformer. *Journal of Machine Learning Research*, 21(140):1–67.

Faisal Rahutomo, Teruaki Kitasuka, and Masayoshi Aritsugi. 2012. Semantic cosine similarity. In *The 7th international student conference on advanced science and technology ICAST*, volume 4, page 1.

Tianxiao Shen, Tao Lei, Regina Barzilay, and Tommi S. Jaakkola. 2017. Style transfer from non-parallel text by cross-alignment. In *Advances in Neural Information Processing Systems 30: Annual Conference on Neural Information Processing Systems 2017*, pages 6830–6841, Long Beach, CA, USA.

Oleh Shliazhko, Alena Fenogenova, Maria Tikhonova, Vladislav Mikhailov, Anastasia Kozlova, and Tatiana Shavrina. 2022. mGPT: Few-Shot Learners Go Multilingual. *CoRR*, abs/2204.07580.

Yuqing Tang, Chau Tran, Xian Li, Peng-Jen Chen, Naman Goyal, Vishrav Chaudhary, Jiatao Gu, and Angela Fan. 2020. Multilingual translation with extensible multilingual pretraining and finetuning. *CoRR*, abs/2008.00401.

Zeerak Waseem and Dirk Hovy. 2016. Hateful symbols or hateful people? predictive features for hate speech detection on Twitter. In *Proceedings of the NAACL Student Research Workshop*, pages 88–93, San Diego, California. Association for Computational Linguistics.

Michael Wiegand and Josef Ruppenhofer. 2021. Exploiting emojis for abusive language detection. In *Proceedings of the 16th Conference of the European Chapter of the Association for Computational Linguistics: Main Volume*, pages 369–380, Online. Association for Computational Linguistics.

Thomas Wolf, Lysandre Debut, Victor Sanh, Julien Chaumond, Clement Delangue, Anthony Moi, Pierric Cistac, Tim Rault, Rémi Louf, Morgan Funtowicz, Joe Davison, Sam Shleifer, Patrick von Platen, Clara Ma, Yacine Jernite, Julien Plu, Canwen Xu, Teven Le Scao, Sylvain Gugger, Mariama Drame, Quentin Lhoest, and Alexander M. Rush. 2020. Transformers: State-of-the-art natural language processing. In *Proceedings of the 2020 Conference on Empirical Methods in Natural Language Processing: System Demonstrations, EMNLP 2020 - Demos, Online, November 16-20, 2020*, pages 38–45. Association for Computational Linguistics.

Linting Xue, Noah Constant, Adam Roberts, Mihir Kale, Rami Al-Rfou, Aditya Siddhant, Aditya Barua, and Colin Raffel. 2021. mT5: A massively multilingual pre-trained text-to-text transformer. In *Proceedings of the 2021 Conference of the North American Chapter of the Association for Computational Linguistics: Human Language Technologies*, pages 483–498, Online. Association for Computational Linguistics.

Marcos Zampieri, Preslav Nakov, Sara Rosenthal, Pepa Atanasova, Georgi Karadzhov, Hamdy Mubarak, Leon Derczynski, Zeses Pitenis, and Çağrı Çöltekin. 2020. SemEval-2020 task 12: Multilingual offensive language identification in social media (OffensEval 2020). In *Proceedings of the Fourteenth Workshop on Semantic Evaluation*, pages 1425–1447, Barcelona (online). International Committee for Computational Linguistics.

Junbo Jake Zhao, Yoon Kim, Kelly Zhang, Alexander M. Rush, and Yann LeCun. 2018. Adversarially regularized autoencoders. In *Proceedings of the 35th International Conference on Machine Learning, ICML 2018*, volume 80 of *Proceedings of Machine Learning Research*, pages 5897–5906, Stockholm, Sweden. PMLR.